\documentclass{llncs}
\usepackage[utf8]{inputenc}
\usepackage[T1]{fontenc}
\usepackage{llncsdoc}
\usepackage{graphicx}
\usepackage[export]{adjustbox}
\usepackage{url}

\begin{document}
\pagenumbering{gobble}
\newcounter{save}\setcounter{save}{\value{section}}
{\def\addtocontents#1#2{}%
\def\addcontentsline#1#2#3{}%
\def\markboth#1#2{}%
\title{Gender Prediction in English-Hindi Code-Mixed Social Media Content : Corpus and Baseline System}
\author{\textbf{Ankush Khandelwal, Sahil Swami, Syed Sarfaraz Akhtar} and \textbf{Manish Shrivastava}}
\institute{ Language Technologies Research Centre, International Institute of Information
Technology, Hyderabad }

\maketitle
\begin{abstract}
The rapid expansion in the usage of social media networking sites leads to a huge amount of unprocessed user generated data which can be used for text mining. Author profiling is the problem of automatically determining profiling aspects like the author's gender and age group through a text is gaining much popularity in computational linguistics. Most of the past research in author profiling is concentrated on English texts  \cite{1,2}. However many users often change the language while posting on social media which is called code-mixing, and it develops some challenges in the field of text classification and author profiling like variations in spelling, non-grammatical structure and transliteration \cite{3}. There are very few English-Hindi code-mixed annotated datasets of social media content present online \cite{4}. In this paper, we analyze the task of author's gender prediction in code-mixed content and present a corpus of English-Hindi texts collected from Twitter which is annotated with author's gender. We also explore language identification of every word in this corpus. We present a supervised classification baseline system which uses various machine learning algorithms to identify the gender of an author using a text, based on character and word level features.
\begin{keywords}
author profiling, code-mixing, language detection, linguistics, SVM, random forest
\end{keywords}
\end{abstract}

\section{Introduction}
Author profiling is the procedure of identifying profiling aspects such as gender, age group, native language or personality type by studying how the language is shared by people. With the advancement of research in natural language processing and computational linguistics over the past years, it is the problem of growing importance due to their diverse applications in online user reviews, advertisements, opinion mining, sentiment analysis, forensics etc. The automatic processing of social media content has gained a lot of interest in the recent times because of the high importance of such data for linguistic analysis. With huge amount of social media data readily available online, there has been some remarkable research in the field of author profiling \cite{1,5}, but most of it is concentrated on English texts. Users often switch between language while posting on social media \cite{4}. For example, consider two tweets, \textit{``Congress jaisi party jo votes k liye kabhi triple talaq ka mudda avoid krti thi wo bhi wah wahi batorne ne padi hai \#wah\_re\_politics''} (A party like Congress which used to subtly issue a triple divorce issue for votes, has also got to collect it, \#wow\_politics) and \textit{``Jab koi baat bigad jaye, jab koi mushkil pad jaye..... Tum dena saath mera, o humnavaaz.....!!''} (
When something goes wrong, when there is a problem ... please be with me, oh my love ..... !!) . The first tweet contains words both in English and Hindi and all the words in Hindi are transliterated to Latin script. Such insertion of words, phrases, and morphemes of one language into a statement or an expression of another language is called \textit{code-mixing} \cite{6}. The latter tweet in which every word in Hindi is just transliterated to Latin script, is an example of \textit{code-switching} \cite{7}. Although both the terms are well known in multilingual linguistics research domain \cite{8,9}, in rest of the paper we will take it as code-mixing for denoting both cases.\\ Social networking platforms like Facebook and Twitter are the reason behind the increment of code-mixing in texts which was typically perceived in the spoken language \cite{8,9}. \cite{4} provides a greater view of how often multilingual speakers switch between languages on social media and provides new challenges because of the variations in spellings, transliterations and non-grammatical structure. \\ 
This paper specifically aims at building the annotated dataset for gender prediction. In the later part of the paper, we analyze the task of gender prediction in English-Hindi code-mixed social media content. We describe and present for the first time a freely available annotated corpus containing English-Hindi code-mixed tweets from Twitter. The tweets are annotated with gender tags \textit{(\textbf{M}ale/\textbf{F}emale)} and each words is annotated with Language tags \textit{(\textbf{Hi}ndi/ \\\textbf{En}glish/\textbf{O}ther)}, for example, \textit{triple} is an English word, \textit{mudda} is a Hindi word and all the proper nouns, symbols, punctuations, emojis, urls, hashtags, mentions and special symbols are annotated as \textit{Others}. \\
The structure of the paper is as follows, we begin by describing the
requirement for code-mixed dataset in Section 2. Section 3 contains the description of corpus and the annotation scheme. Section 4 summarizes our supervised classification system which includes pre-processing of the corpus and the feature extraction followed by the method used to predict the gender. In the next subsection, we describe the classification model and the results of the experiments conducted using character and word level features. In the last section, we conclude the paper followed by future work and references.
}

\section{Requirement of Code-mixed Dataset for Gender Prediction}
Gender prediction has lot of applications like  customer review, personalization and customization, public-opinion management, online ads placement, purchase planning, detection of inflammatory text and cyber-bullying. People in India often mix English and Hindi (Hindi words transliterated to Latin script) while posting on social media. Hence there is a need to develop a dataset for predicting gender in English-Hindi code-mixed texts. Although there are many English datasets present on author profiling \cite{1,2,19}, this is the first attempt at building a English-Hindi code-mixed dataset. The features used in the classification model are based on character and word level. 

\section{Dataset and Annotation Scheme}

In this section, we explain the technique used in the creation and annotation of the corpus.

\subsection{Data Collection and Statistics}

We use the python package twitterscraper\footnote{https://github.com/taspinar/twitterscraper}  to scrap tweets from twitter which uses the advance search option of twitter. We mined the tweets for the past two years involving some social and political issues which have been prevalent in India recently. The tweets were collected from 1000 twitter accounts. The statistics are given in Table 1. The corpus comprises of tweets mined using the keywords like \textit{notebandi (demonetization)}, \textit{triple talaq}, \textit{general service tax (GST), surgical strike, rape} etc.\\ Tweets extracted are stored in json format which consists of all the information about a tweet like timestamp of the tweet, tweet id, tweet text, user name and full name of author and also retweets and replies as shown in Figure 1. Two annotators were involved in the annotation process, both of them being native Hindi speakers and took around 50 hours to complete the annotation. Annotators were in complete agreement with each other to identify the gender of the tweet. The Final Statistics of the dataset are given in Table 2. \\

\begin{figure}[!h]
\begin{center}
\includegraphics[scale=0.5]{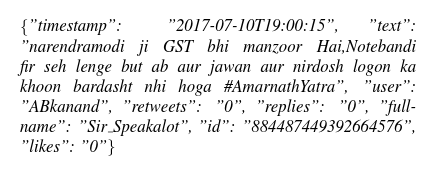} 
\caption{Json format of a tweet.}
\label{fig.1}
\end{center}
\end{figure}

\setlength{\tabcolsep}{8.5pt}
\renewcommand{\arraystretch}{1}
\begin{table}
\begin{center}
\caption[1]{Statistics of the tweets collected.}
\begin{tabular}{{cc}}

      \hline
      \textbf{} & \textbf{Number}\\
      \hline
       Total tweets collected & 11,280 \\
      \hline
       Tweets in English & 4140 \\
      \hline
      Tweets in Hindi & 3245\\
      \hline
      Code-Mixed tweets & 4155\\
      \hline
       Annotators & 2 \\
      \hline
       Human Hours devoted for Annotation (each) & 50\\
      \hline
       Tweets annotated using profile picture for unisex name & 800\\
      \hline
       Tweets annotated using linguistic details \\ for missing profile picture for unisex name& 600\\
      \hline
       Tweets identified for fake accounts & 1280\\
      \hline
       Tweets annotated for alphanumeric names & 445\\
      \hline

\end{tabular}
\end{center}
\end{table}

\subsection{Gender Annotation}

First step is to label the tweets with gender tags. Twitter does not store the gender of users during signup, hence it can not be extracted from user’s profile page. Hence the tags have to be assign manually. There are various factors deciding the gender of an author for manual tagging. First name and username often provide most of the insights in deciding the gender of a tweeter. It is possible to identify the gender of people through their real names like ``Akshay'', ``Shekhar'' and ``Raj'' are male names, while ``Shalini'', ``Rekha'' and ``Parul'' are female names. But there are some exceptions to this identification through names, in case of unisex names like ``Kiran'', ``Manpreet'', we actually have to go through their twitter profile manually using the username and assign the gender by looking at the profile pic of the user. If there is no profile pic then we look for any linguistic details in the tweet such as presence of words that is specific to a gender in both language, for example, Hindi words like ``likhungi'', ``karungi'' and English words like ``my husband'' are specific to females. Same process is used for those users who provide fake accounts or made up username or full name which are other than their real names and also usernames consisting of combination of alphanumeric and special characters, like ``shaktimaan'', ``therealjoker\$\$'' etc. Finally we annotated \textit{4015} code-mixed tweets containing \textit{2029} tweets written by males and \textit{1986} tweets written by females. We have included similar number of tweets from both male and female authors to make the classification models more robust \cite{20}.  
 
\setlength{\tabcolsep}{8.5pt}
\renewcommand{\arraystretch}{1}
\begin{table}
\begin{center}
\caption[1]{Statistics of final corpus}
\begin{tabular}{{cc}}
      \hline
      \textbf{} & \textbf{Number}\\
      \hline
       Tweets included in final corpus & 4015 \\
      \hline
       Total words  & 75,000\\
      \hline
       Words in Hindi (Transliterated to Latin script) & 43,349\\
      \hline
       Words in English & 31,545 \\
      \hline
       Others (Emojis,Punctuations, etc) & 106 \\
      \hline
       Male Tweets & 2029\\
      \hline
       Female Tweets & 1986\\
      \hline

\end{tabular}
\end{center}
\end{table}

\subsection{Language Annotation}

  Native speakers of Hindi and proficient in English, labelled the source language of the  words in the tweets. Three kind of tags are given, \textbf{En} for words present in English vocabulary like ``family'' and ``Children''. Second is \textbf{Hi} for words present in Hindi vocabulary but transliterated to Latin script like ``samay'' (time) and ``aamaadmi'' (common man). Rest of the tokens which includes proper nouns, numbers, dates, urls, hashtags, mentions, emojis and punctuations are assigned \textbf{O}(other) tag. Some words are common in both Hindi and English (ambiguous words), like `to'(`but' in Hindi) and `is'(`This' in hindi) , for this scenario annotators understood the context of the tweet and based on that the words were being annotated. For example, consider two sentences `this is an umbrella' and `Is bar Modi Sarkar'. So `is' in first sentence is in English and it is transliterated to Latin script in latter sentence.

\begin{figure}[!h]
\begin{center}
\includegraphics[scale=0.5]{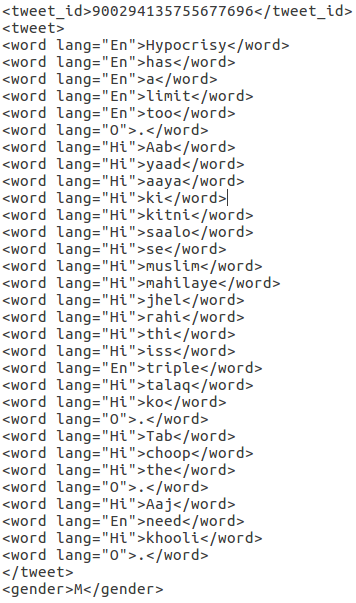} 
\caption{An annotated tweet.}
\label{fig.2}
\end{center}
\end{figure}
After doing gender annotations and language annotations, usernames and full names are removed from the corpus. Figure 2 describes the annotation scheme through an example annotation.  Each annotation starts with tweet id. Each tweet is enclosed within \textless tweet\textgreater \textless /tweet\textgreater. Every line within these tags contains words in the tweet which are enclosed within \textless word lang=`` ''\textgreater\textless /word\textgreater~containing the language annotation for each word. Every line within these tags contains a token with the language tag separated by a space. Last line contains the gender tag assigned to the tweet enclosed withing \textless gender\textgreater \textless /gender\textgreater~tags. 

\subsection{Error Analysis}
Social media users often make spelling mistakes while posting a text. We replaced all such typos with their correct version. For words in Hindi that were transliterated to Latin script, we adopted a common spelling for all those words across the dataset. For example, `dis' is often used as a short form for `this', so we replaced every occurrence of `dis' to `this' in the corpus.

\section{System Architecture}

In this section, we describe our machine learning model which is trained and tested on the corpus described in the previous section.

\subsection{Preprocessing of Corpus}

Tokenization is the first step in preprocessing in which the words in the tweets are separated using space as the delimiter followed by converting words to lower cases. Then, the punctuation marks are removed from the tokenized tweet. All the hashtags, mentions and urls, are stored and converted to `hashtag', `mention' and `url' respectively. Almost every tweet contains hashtags which can be included in the tokenized words, hence we segregated the hashtags which consists of words combined with camel cases \textit{(hashtag decomposition)}\cite{11,12}. Finally the tokenized tweets are stored along with the gender as the target class.

\subsection{Classification Features}

Following is the description of the features used to build attribute vectors for training our classification model \footnote{Threshold values described are taken after empirical fine tuning}. Character and word level features are used for the classification \cite{12}. 

\subsubsection{Character N-grams}

We have included this feature in our system as it is language independent and does not require any pre-processing and previous knowledge like tokenization, stemming and stopwords removal. Thus, it can be an advantage in case of classification of code-mixed tweets\cite{13,14}.
Since the number of all existing n grams is very large, we downsample them using their frequency. Only those n grams are taken which occur at least ten times in the corpus which in turn will reduce the size of the feature vector. 

\subsubsection{Bag of words}

Bag of words have proven to be a less important feature feature for text classification than Character grams in case of text classification \cite{14}. However we take this as a feature in our experiments to analyze its performance in case of gender prediction. We take into account the word n grams where n varies from 1 to 3. In this case, we take only those n-grams which occur at least ten time in the corpus which reduce the less important and noisy n grams.

\subsubsection{Reference tokens}

We identified the tokens which occurs for more than 60\% in a gender and occur more than five times in the corpus set and took them as a features for the classification models . According to \cite{15}, there are some tokens which distinguish between males and females , thus storing indicative tokens can provide better classification. We calculated the value for each token as ratio of the frequency of the token which belongs to a class and total frequency of the token in the corpus. We made separate dictionaries for Hindi and English words and took the reference tokens from each dictionary.
For gender prediction , only those tokens are taken as features for classification which have a score \textgreater = 0.6 and occur at least two times in the training corpus.

\setlength{\tabcolsep}{8.5pt}
\renewcommand{\arraystretch}{1}
\begin{table}
\begin{center}
\caption[1]{Accuracy of each feature using naive bayes classifier}
\begin{tabular}{{cc}}

      \hline
      \textbf{Features (in \%)} & \textbf{Naive Bayes}\\
      \hline
      Character N grams & 87.3 \\
      \hline
      Bag-of-words & 78.3 \\
      \hline
     Reference Tokens & 71\\
      \hline
       Top Hashtags & 54.5\\
      \hline
       All features & \textbf{85}\\
      \hline

\end{tabular}
\end{center}
\end{table}

\setlength{\tabcolsep}{8.5pt}
\renewcommand{\arraystretch}{1}
\begin{table}
\begin{center}
\caption[2]{Accuracy of each feature using kernel support vector machines}
\begin{tabular}{{cc}}

     \hline
      \textbf{Features (in \%)} & \textbf{Kernel SVM}\\
      \hline
      Character N grams & 89.7\\
      \hline
      Bag-of-words & 83.6\\
      \hline
     Reference Tokens & 87.5\\
      \hline
       Top Hashtags & 56.4\\
     
      \hline
       All features & \textbf{89.5}\\
      \hline

\end{tabular}
\end{center}
\end{table}

\setlength{\tabcolsep}{8.5pt}
\renewcommand{\arraystretch}{1}
\begin{table}
\begin{center}
\caption[1]{Accuracy of each feature using random forest classifier}
\begin{tabular}{{cc}}

      \hline
      \textbf{Features (in \%)} & \textbf{Random Forest Classifier}\\
      \hline
      Character N grams & 85.6\\
      \hline
      Bag-of-words & 84.5\\
      \hline
     Reference Tokens & 85.8\\
      \hline
       Top Hashtags & 54.6\\
     
      \hline
       All features & \textbf{88.4}\\
      \hline

\end{tabular}
\end{center}
\end{table}
\subsection{Classification Approach and Results}

Previous researches have shown that for text classification and sentiment analysis, support vector machines and random forest classifiers provide better results than rest of the machine learning models \cite{16,17}. Since the size of feature vectors formed are very large, we applied chi square feature selection  algorithm which reduces the size of our feature vector to 1000.  In our system, we have used SVM with rbf kernel as they perform efficiently in case of high dimensional feature vectors. For training our system classifier, we have used Scikit-learn \cite{18}.\\
We experimented with three different classifiers, namely, SVM with radial basis function kernel, random forest and naive bayes classifier. We train the classifiers with two different scenarios, in one case feature vectors are formed based on Hindi words and other on English words in a tweet and later combined to train the classifiers but they did not perform well as compared to language independent feature vectors in the second case.\\
We performed 10-fold cross validation on 4015 code-mixed tweets by dividing the corpus in ten equal parts with nine parts as training corpus and one for testing. Since tweets were collected from small number of users, we ensured that all tweets from an user must occur either in training data or test data so that gender classification should not be based on specific words used by that user multiple times. Finally, the mean of accuracy of each iteration is taken as the final accuracy of the classification model. Table 1 describes the accuracy for each feature when trained using different classifiers.

All the experiments are carried out by performing grid search on every classification model.
After performing all experiments, we observed that character n grams performed better in all classification model and gives a highest accuracy of 89.7\% with kernel SVM. Random forest classifier performs the worst in case of character n grams but it outperforms naive bayes classifiers when all features are taken during the training of classifier. The best performance is given by SVM with radial basis functions kernel and gives an accuracy of 89.5 \%.

%

\section{Conclusion and Future Work}
In this work, we introduce a freely available dataset for gender detection for code-mixed texts. The dataset consists of 4015 English-Hindi code-mixed tweets annotated with gender and language tags. We find some interesting information after annotating the dataset.  Females tends to use more punctuations and hashtags than males. The average number of hashtags in a female tweet in our dataset is two as compared to one in the case of males. Similarly, females use an average of three punctuations in a tweet compared to one in case of males. The number of punctuations and hashtags in a tweet can be included in classification features to improve the classification. On the basis of tweets collected, the average number of words in a tweet by female is twenty which is same as that for a tweet by male, hence it is not useful to use number of words as a feature in the classification. We included tweets on social and political issues in India, but this dataset can be extended to include tweets on various other topics like sports and entertainment. Furthermore, we plan to annotate the dataset with part-of-speech (POS) tags which should help in understanding the structure of code-mixed sentences and can yield better results when used as features for classification. The annotations and experiments described in this paper can also be carried out for code-mixed texts containing more than two languages from multilingual societies in future also it will be interesting to use neural network for classification in the future experiments.  Comparing training with code-mixed tweets with training with a merged dataset of monolingual tweets in English and Hindi could be an interesting future work.



\begin{thebibliography}{[MT1]}

\bibitem[1]{1}
Estival, Dominique, et al. "Author profiling for English emails." Proceedings of the 10th Conference of the Pacific Association for Computational Linguistics. 2007.

\bibitem[2]{2}
Peersman, Claudia, Walter Daelemans, and Leona Van Vaerenbergh. "Predicting age and gender in online social networks." Proceedings of the 3rd international workshop on Search and mining user-generated contents. ACM, 2011.

\bibitem[3]{3}
Barman, Utsab, et al. "Code mixing: A challenge for language identification in the language of social media." Proceedings of The First Workshop on Computational Approaches to Code Switching. 2014.

\bibitem[4]{4}
Vyas, Yogarshi, et al. "POS Tagging of English-Hindi Code-Mixed Social Media Content." EMNLP. Vol. 14. 2014.

\bibitem[5]{5}
Marquardt, James, et al. "Age and gender identification in social media." Proceedings of CLEF 2014 Evaluation Labs. 2014.

\bibitem[6]{6}
Myers-Scotton, Carol. Duelling languages: Grammatical structure in codeswitching. Oxford University Press, 1997.

\bibitem[7]{7}
Gumperz, John J. Discourse strategies. Vol. 1. Cambridge University Press, 1982.
\bibitem[8]{8}
Danet, Brenda, and Susan C. Herring, eds. The multilingual Internet: Language, culture, and communication online. Oxford University Press on Demand, 2007.
\bibitem[9]{9}
Cárdenas-Claros, Mónica Stella, and Neny Isharyanti. "Code-switching and code-mixing in internet chatting: Between’yes,” ya,’and’si’-a case study." The Jalt Call Journal 5.3 (2009): 67-78.

\bibitem[10]{10}
Auer, Peter, and Raihan Muhamedova. "Embedded language’and ‘matrix language’in insertional language mixing: Some problematic cases." Rivista di linguistica 17.1 (2005): 35-54.

\bibitem[11]{11}
Belainine, Billal, Alexsandro Fonseca, and Fatiha Sadat. "Named Entity Recognition and Hashtag Decomposition to Improve the Classification of Tweets." Proceedings of the 2nd Workshop on Noisy User-generated Text (WNUT). 2016.

\bibitem[12]{12}
Khandelwal, A., Swami, S., Akhtar, S.S., Shrivastava, M.: Classification Of Spanish
Election Tweets (COSET) 2017: Classifying Tweets using Character and Word
Level Features. In: Proceedings of the Second Workshop on Evaluation of Human
Language Technologies for Iberian Languages (IberEval 2017). CEUR Workshop
Proceedings. CEUR-WS.org, Murcia (Spain) (September 19 2017)
\bibitem[13]{13}
Lodhi, Huma, et al. "Text classification using string kernels." Journal of Machine Learning Research 2.Feb (2002): 419-444.
\bibitem[14]{14}
Cavnar, William B., and John M. Trenkle. "N-gram-based text categorization." Ann Arbor MI 48113.2 (1994): 161-175.
\bibitem[15]{15}
Argamon, Shlomo, et al. "Gender, genre, and writing style in formal written texts." TEXT-THE HAGUE THEN AMSTERDAM THEN BERLIN- 23.3 (2003): 321-346.
\bibitem[16]{16}
Joachims, Thorsten. "Transductive inference for text classification using support vector machines." ICML. Vol. 99. 1999.
\bibitem[17]{17}
Mccord, Michael, and M. Chuah. "Spam detection on twitter using traditional classifiers." international conference on Autonomic and trusted computing. Springer, Berlin, Heidelberg, 2011.
\bibitem[18]{18}
Pedregosa, Fabian, et al. "Scikit-learn: Machine learning in Python." Journal of Machine Learning Research 12.Oct (2011): 2825-2830.
\bibitem[19]{19}
Lusa, Lara, et al. "Challenges in projecting clustering results across gene expression–profiling datasets." JNCI: Journal of the National Cancer Institute 99.22 (2007): 1715-1723.
\bibitem[20]{20}
Du, Mian, et al. "Supervised classification using balanced training." International Conference on Statistical Language and Speech Processing. Springer, Cham, 2014.
\end{thebibliography}
\end{document}